\documentclass[conference]{IEEEtran}
\usepackage{fancyhdr, epsfig, amsthm, amsmath, amssymb, amsfonts, subfigure,color,dsfont,float}

\usepackage{cite}
\usepackage{algorithmic}
\usepackage{graphicx}
\usepackage{textcomp}
\usepackage{dblfloatfix} 
\usepackage[table,xcdraw]{xcolor}
\usepackage{tabularx}
\usepackage{tikz}
\usepackage{enumitem}
\usepackage{url}
\usepackage{booktabs}
\usepackage{array}
\usepackage{multirow}
\usepackage{booktabs} 
\usepackage{siunitx}  

\usepackage{xfrac}
\usepackage{nicefrac}
\usepackage{psfrag}
\usepackage{times}
\usepackage{balance} 
\usepackage{amssymb}
\usepackage{amsfonts}
\usepackage{bm}
\usepackage[bottom]{footmisc}
\usepackage{mwe}
\usepackage{color}
\usepackage{booktabs,multirow}
\usetikzlibrary{patterns,arrows}

\definecolor{color1}{RGB}{237, 191, 193}
\definecolor{color2}{RGB}{229, 153, 157}
\definecolor{color3}{RGB}{225, 123, 116}
\definecolor{color4}{RGB}{217,87,77}
\definecolor{color5}{RGB}{203,52,38}

\def\l{\left}
\def\r{\right}
\def\({\l(}
\def\){\r)}
\def\[{\l[}
\def\]{\r]}

\def\BibTeX{{\rm B\kern-.05em{\sc i\kern-.025em b}\kern-.08em
    T\kern-.1667em\lower.7ex\hbox{E}\kern-.125emX}}

\begin{document}

\title{Split Learning Meets Koopman Theory for \\Wireless Remote Monitoring and Prediction}


\author{
\IEEEauthorblockN{Abanoub M. Girgis\IEEEauthorrefmark{1}, Hyowoon Seo\IEEEauthorrefmark{1}, Jihong Park\IEEEauthorrefmark{2}, Mehdi Bennis\IEEEauthorrefmark{1}, and Jinho Choi\IEEEauthorrefmark{2}}
\IEEEauthorblockA{
\IEEEauthorrefmark{1}Centre for Wireless Communications\\
University of Oulu, Oulu 90014, Finland\\
Email: \{abanoub.pipaoy,\;hyowoon.seo,\;mehdi.bennis\}@oulu.fi\\
\IEEEauthorrefmark{2}School of Information Technology\\
Deakin University, Geelong, VIC 3220, Australia\\
Email: \{jihong.park,\;jinho.choi\}@deakin.edu.au
}}

\maketitle 

\begin{abstract}
  Remote state monitoring over wireless is envisaged to play a pivotal role in enabling beyond 5G applications ranging from remote drone control to remote surgery. One key challenge is to identify the system dynamics that is non-linear with a large dimensional state. To obviate this issue, in this article we propose to train an autoencoder whose encoder and decoder are split and stored at a state sensor and its remote observer, respectively. This autoencoder not only decreases the remote monitoring payload size by reducing the state representation dimension, but also learns the system dynamics by lifting it via a Koopman operator, thereby allowing the observer to locally predict future states after training convergence. Numerical results under a non-linear cart-pole environment demonstrate that the proposed split learning of a Koopman autoencoder can locally predict future states, and the prediction accuracy increases with the representation dimension and transmission power.

\end{abstract}


\section{Introduction}
\label{Introduction}

Monitoring system states at remote locations with high accuracy is crucial in a variety of  fifth generation (5G) emerging applications and beyond~\cite{park2020extreme,bennis2018ultrareliable}. Remote drone control \cite{9013181} is one example, wherein each control command is determined in real time by monitoring sensor states such as the drone's position, velocity, and temperature. There are several high-stake use cases including smart factory automation \cite{liu2019taming}, driverless cars \cite{zeng2019joint}, remote surgery \cite{meng2004remote}, and bushfire detection \cite{liu2010smart} to mention a few, all of which mandate accurate system state acquisition with extremely low latency. For remote monitoring, wireless connectivity is essential, yet it comes at the cost of incurring distorted state observations due to the channel noise under uncoded analog transmissions (or non-negligible latency under digital transmissions) \cite{vazquez2014analog,hassanin2013analog}. One may correct distortion by understanding the system dynamics and comparing it with the observations. Unfortunately, system dynamics is often non-linear and high-dimensional, which poses a bigger challenge to solve.

To fill this void, in this article we propose a novel deep learning framework for wireless remote monitoring and prediction, inspired by Koopman autoencoder \cite{lusch2018deep} and split learning \cite{Vepakomma:2018:Splita,park2019wireless}. To illustrate, suppose a non-linear state recurrence relation $\mathbf{x}_{t+1}= \mathbf{f}(\mathbf{x}_{t})$ for a remote sensor state $\mathbf{x}_{t}$ at time~$t$ with an unknown non-linear function $\mathbf{f}(\cdot)$. According to Koopman operator theory~\cite{koopman1931hamiltonian}, such non-linear system dynamics can be transformed into linear system dynamics using a Koopman operator $\mathcal{K}$ and its associated $q$ eigenfunctions. A prior work \cite{lusch2018deep} has shown that $\mathcal{K}$ (or its matrix representation~$\mathbf{K}$ which will be discussed later on) is trainable with high accuracy using an autoencoder neural network architecture. Motivated by this, as illustrated in Fig.~1, we consider a tripartite autoencoder that consists of: (i) an encoder $\Psi$ of the input state $\mathbf{x}_{t}$, (ii) two dense hidden layers with width $q$, and (iii) a decoder $\Psi^{-1}$. While the final output $\mathbf{x}_{t+1}$ after decoding is predicted by feeding the input $\mathbf{x}_t$ to the encoder, the Koopman operator $\mathcal{K}$ is identified through the hidden layers. This end-to-end autoencoder is split into two parts, such that (i) is locally stored at the sensor, while (i) and (ii) are offloaded to its remote observer. Then, by training the autoencoder end-to-end, $\mathcal{K}$ is learned by the remote observer. 

\begin{figure*}[htbp]
\centering
\subfigure[Split Koopman autoencoder architecture.]{\includegraphics[width=.57\linewidth]{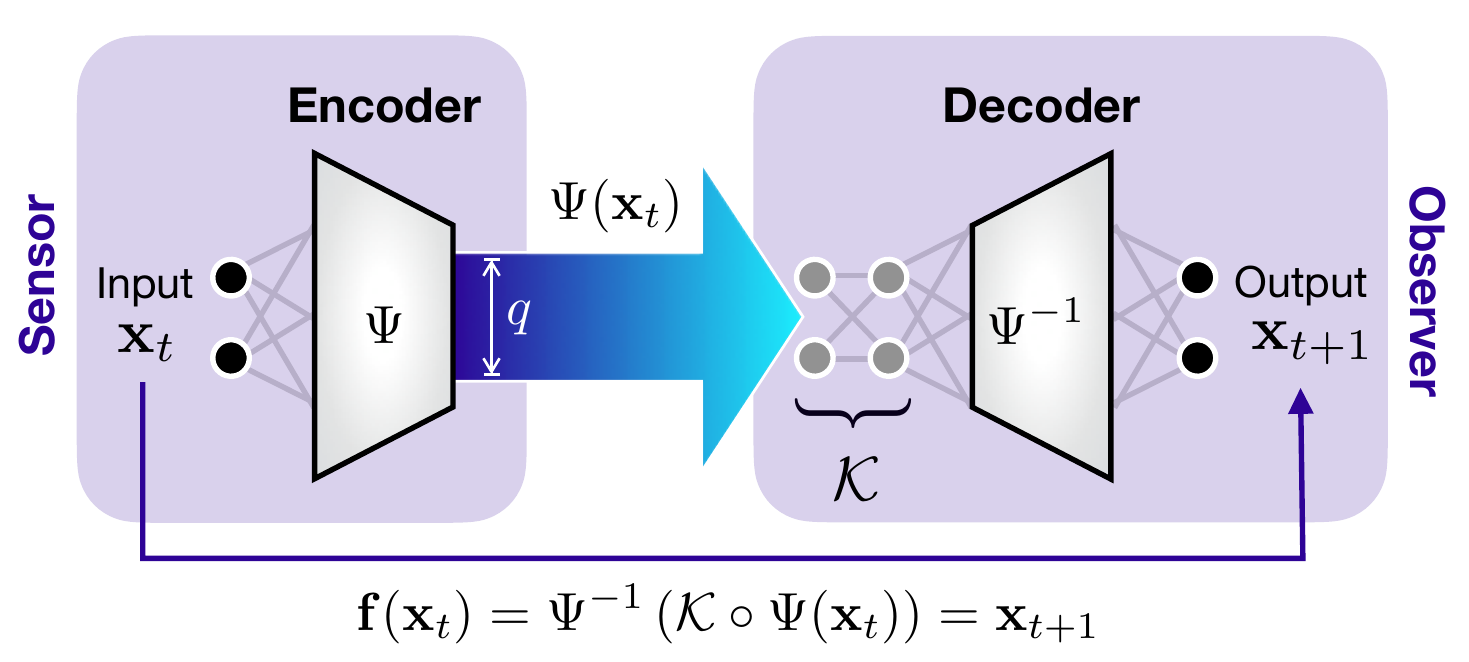}} \hspace{20pt}
\subfigure[Koopman approximation.]{\includegraphics[width=.37\linewidth]{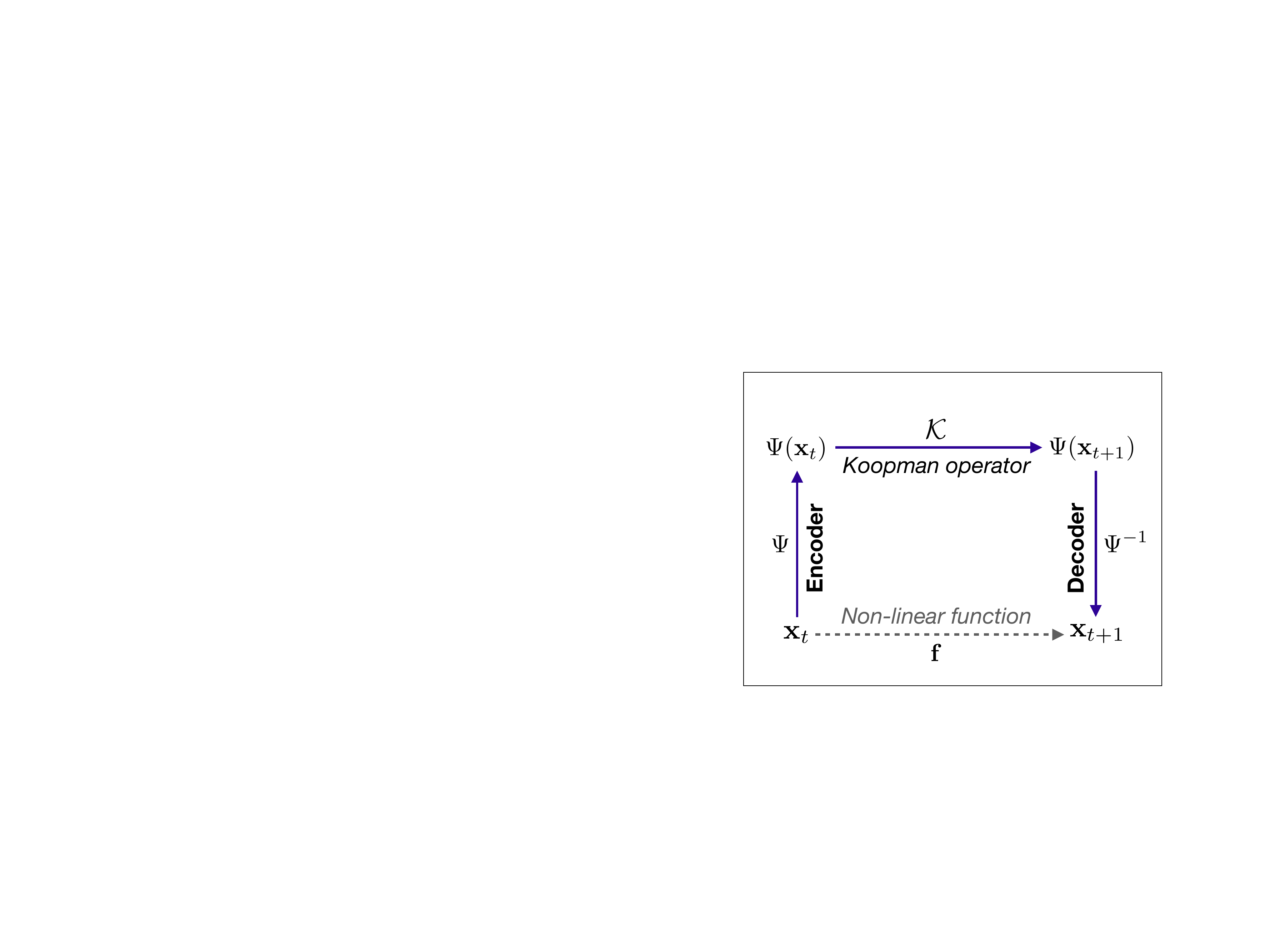}}
\caption{Schematic illustrations of (a) the split Koopman autoencoder and (b) the approximation of the non-linear function $\mathbf{f}$ via the Koopman operator $\mathcal{K}$.} 
\label{fig_System model}
\end{figure*}

The aforementioned operations entail two-way iterative communications during which the sensor transmits the encoder output over a wireless channel, while the observer sends back the error feedback of the decoded output, i.e., gradient. After training convergence, the observer understands the non-linear system dynamics through the learned $\mathcal{K}$, and it is capable of locally predicting future state $\mathbf{x}_{t'}$ for any $t'>t$ using past observation $\mathbf{x}_t$. Furthermore, the Koopman autoencoder can recognize the true system dynamics even under non-negligible distortion by the channel noise. To show the feasibility of the proposed split learning of Koopman autoencoder, a remote observer aims to learn the system dynamics of an inverted cart-pole system whose inverted pendulum angle and cart states are non-linear \cite{morasso2019stabilization}. Simulation results validate that the observer can learn the Koopman operator with high accuracy even under a moderate level of channel noise. Moreover, the results show that increasing the autoencoder hidden layer width $q$ (i.e., the number $q$ of Koopman eigenfunctions) and/or transmission power can improve the future state prediction accuracy, highlighting the importance of co-designing deep learning, control, and communication operations.

\vspace{3pt}\noindent\textbf{Related Works.}\quad 
To make a non-linear dynamical system amenable, Jacobian linearization method is a well-known method, providing a linear approximation around the equilibrium point based on Taylor series expansion~\cite{schei1997finite}. By nature, the approximation becomes vacuous when it goes far from the equilibrium point. Koopman operator theory~\cite{koopman1931hamiltonian} suggests an alternative solution via lifting up the finite-dimensional non-linear dynamical system to an infinite-dimensional linear system. A recent work~\cite{lusch2018deep} makes the infiniteness tractable by training a finite-dimensional autoencoder. Nevertheless, the training is offline and standalone while assuming noise-free observables. This limitation questions its feasibility for real-time remote monitoring, motivating this work.

Meanwhile, remote system monitoring/controlling scenarios have long been studied in the context of wireless communications. Particularly, a recent work~\cite{eisen2019control} proposes a communication scheduling method reflecting control system dynamics. In our recent work~\cite{girgis2021predictive}, while running such a control dynamics aware scheduler, the system locally predicts unscheduled state and control messages via Gaussian process regression (GPR), improving the overall system control stability under limited wireless resources. Nonetheless, these works commonly assume linear system dynamics that is often not realistic.

\section{Problem Statement}
\label{System_Model}
Consider a system composed of a sensor that measures the state information of a non-linear dynamical system and an observer that obtains such information from the sensor by communicating over a wireless link, as illustrated in Fig.~\ref{fig_System model}. The $D$-dimensional state of the non-linear dynamical system at discrete-time $t \geq 0$ is denoted by vector $\mathbf{x}_{t} \in \mathbb{R}^{D}$, and its state transition is described as
\begin{align}\label{eq_autonomous_fn}
\mathbf{x}_{t+1} =  \mathbf{f} ( \mathbf{x}_{t}, \mathbf{n}_{s,t} ), 
\end{align}
where $\mathbf{n}_{s,t} \in \mathbb{R}^{D}$ is a $D$-dimensional random system noise vector with covariance matrix $\mathbb{E}[\mathbf{n}_{s,t}^T \mathbf{n}_{s,t} ] = N_{s}\mathbf{I}_D$. Note that $\mathbf{f}(\cdot)$ is the non-linear state transition mapping function that drives the dynamical system state forward in time.

The main goal is to keep the system states updated and monitored in real-time at the observer. One naive option is to transmit the system states at every time instant from the sensor to the observer. Supposing both the sensor and observer are equipped with a single antenna, we consider a time-division approach. For a given measured system state $\mathbf{x}_t$, the sensor transmits its states over $D$ orthogonal resources, and the received signal $\mathbf{y}_t$ at the remote observer is represented as
\begin{align}\label{eq_Received_Signal}
\mathbf{y}_t = \sqrt{P} (\mathbf{h}_t \ast \mathbf{x}_t) + \mathbf{n}_{c,t},
\end{align}
where $P$ is the transmission power scaling for the signal, $\mathbf{h}_{t} \in \mathbb{R}^{D}$ is the wireless channel vector corresponding to the transmit signal $\mathbf{x}_t$, incorporating the effect of small scale Rayleigh fading and path-loss large scale fading, $\ast$ is the element-wise multiplication operator, and $\mathbf{n}_{c,t}$ is the additive white Gaussian noise vector of the channel with covariance matrix $\mathbb{E}[\mathbf{n}^{T}_{c,t} \mathbf{n}_{c,t}] = N_{c} \mathbf{I}_{D}$.
The wireless channel is modeled as a block fading. For estimating the original states at the receiver, we assume the channel state information is given and the received signal is divided by the channel state. The distorted signal of the original system state information $\mathbf{x}_t$ after going through the wireless channel is denoted by $\hat{\mathbf{x}}_t$. Note that in the later sections, we consider transmitting the representation $\mathbf{z}_t$ of the system state $\mathbf{x}_t$ at the sensor and estimate $\mathbf{z}_t$ at the observer.


The problem of real-time system state update when transmitting every measured non-linear dynamical system state from the sensor to the observer is the need of a large amount of communication resources. Instead of real-time transmission, by estimating and predicting future system states will significantly relax the problem while ensuring efficient communication. However, the non-linearity of the dynamical system makes the remote observer hard to estimate the system dynamics and predict the future system states. Hence, we propose a wireless split Koopman autoencoder structure to linearize the non-linear system dynamics and predict the future system states.

\section{Split Learning Koopman Operator}\label{Koopman Operator for non-linear Prediction}
To begin, we revisit the fundamental theory of Koopman operator. Then, we propose a method for applying Koopman operator with split training of the Koopman autoencoder that linearizes the non-linear system dynamics and thereby enables future system state prediction.

\subsection{Basic Theory of Koopman Operator}\label{Koopman Learning}
Typically, a data sample observed from a dynamical system can be seen as a function of system states. Define functions $g:\mathbb{R}^{D}\rightarrow \mathbb{R}$, which we call \emph{observables} of interest and analogous to data samples, which are elements of an infinite-dimensional Hilbert space. The Koopman operator, denoted by $\mathcal{K}$, is defined as a linear composition operator on the space of the observables as~\cite{9304238}
\begin{align}\label{Koopman_fun1}
\mathcal{K} g \left( \mathbf{x}_{t} \right) = g \circ \mathbf{f} \left( \mathbf{x}_{t}\right),
\end{align}
where $\circ$ denotes the function composition. The linearity of the Koopman operator comes from
\begin{align}\label{Koopman_Linearity}
\mathcal{K} \left[ g_{1}   +  g_{2}\right] \left( \mathbf{x}_{t} \right) = \left[ g_{1} +  g_{2} \right] \circ \mathbf{f}  \left( \mathbf{x}_{t}\right),
\end{align}
for two different observables $g_{1} $ and $g_{2}$. In short, when dealing with non-linear dynamical system, the Koopman operator helps to shift the viewpoint from the system state space to the observable space, giving rise to a linear evolution of system dynamics in the space of the observables as~\cite{birkhoff1932recent}
\begin{align}\label{Koopman_fun2}
\mathcal{K}  g ( \mathbf{x}_{t} ) = g ( \mathbf{x}_{t+1} ).
\end{align}
The linearity of the Koopman operator is an advantage in terms of high fidelity, compared to conventional linearization methods around an equilibrium point, which is inaccurate when it is far from the equilibrium point. However, difficulties come from the fact that the Koopman operator acts in the infinite-dimensional space of observables. To overcome the issue of the infiniteness of the operator relies on finding a special set of observables that span an invariant subspace, in which the Koopman operator works.

 For this purpose, eigenfunctions of the Koopman operator are set of functions that spans such an invariant subspace. Note that a discrete Koopman eigenfunction $\psi(\mathbf{x}_t)$ that corresponds to eigenvalue $\lambda$ satisfies~\cite{mezic2017koopman}
 \begin{align}
 \label{Koopman eigenfunctions}
 \psi \left( \mathbf{x}_{t+1} \right) = \mathcal{K} \psi \left( \mathbf{x}_{t} \right) = \lambda \psi \left( \mathbf{x}_{t} \right).
 \end{align}
 Furthermore, any finite set of eigenfunctions will span an invariant subspace. For example, consider $q$ different eigenfunctions $\psi_1,\psi_2, \dots \psi_q$, which span 
 \begin{align}
 \tilde{\psi}(\mathbf{x}_t) = a_1\psi_1(\mathbf{x}_t) + a_2\psi_2(\mathbf{x}_t) + \cdots + a_q\psi_q(\mathbf{x}_t),
 \end{align}
for some parameters $a_1,a_2,\dots,a_q \in \mathbb{R}$ and the Koopman operator acts on the observables space as
 \begin{align}
 \mathcal{K}\tilde{\psi}(\mathbf{x}_t) = \lambda_1 a_1\psi_1(\mathbf{x}_{t}) + \lambda_2 a_2\psi_2(\mathbf{x}_{t}) + \cdots + \lambda_q a_q\psi_q(\mathbf{x}_{t}),
 \end{align}
 with the corresponding eigenvalues $
 \lambda_1,\lambda_2,\dots,\lambda_q$.
 By introducing a finite-dimensional matrix representation $\mathbf{K} \in \mathbb{R}^{q\times q}$ for the given invariant subspace, we obtain a linear representation of the non-linear dynamical systems as~\cite{yeung2019learning,brunton2016koopman}
 \begin{align}\label{Koopman_vectorvalued}
 \Psi(\mathbf{x}_{t+1}) = \mathbf{K}\Psi(\mathbf{x}_{t}),
 \end{align}
 where $\Psi(\cdot)$ is the concatenated vector of the $q$ eigenfunctions. Note that a given current system state $\mathbf{x}_t$, if we know the concatenated eigenfunction $\Psi$, its inverse $\Psi^{-1}$ and the Koopman matrix $\mathbf{K}$, the future system state $\mathbf{x}_{t+1}$ can be retrieved. However, it is difficult to mathematically predict the future non-linear system states, which induces us to solve this problem by applying machine learning methods.

\subsection{Wireless Split Koopman Autoencoder}\label{Data- driven Koopman Invarian subcace}
We consider a tripartite autoencoder neural network architecture that consists of an encoder that is related to the concatenated eigenfunction $\Psi$, two fully connected hidden layers that constitutes the matrix representation of the Koopman $\mathbf{K}$, and the decoder which is the inverse function $\Psi^{-1}$. The autoencoder is split in two parts, where the encoder is located at the sensor (transmitter) side, while the Koopman hidden layers and the decoder are at the observer (receiver) side. After the split Koopman autoencoder is well-trained, if the sensor sends a representation $\mathbf{z}_t = \Psi(\mathbf{x}_t)$ of the measured system state $\mathbf{x}_t$ over the wireless channel, the observer obtains a noisy representation $\hat{\mathbf{z}}_t$. Then, by passing it through the trained Koopman matrix and decoder, the observer obtains the future system state $\mathbf{x}_{t+1} = \Psi^{-1}(\mathbf{K}\hat{\mathbf{z}}_t)$. Moreover, by applying the learned Koopman matrix $\tau$ times over the last received state representation we obtain all future system states as \begin{align}
\mathbf{x}_{t+\tau} = \Psi^{-1} \left(  \mathbf{K}^{\tau}\hat{\mathbf{z}}_t \right),
\end{align}
for all prediction depth $\tau \in \mathbb{R}^{+}$.

Therefore, by leveraging with the split Koopman autoencoder the real-time remote monitoring of the system consists of two phases. In the first phase, the observer receives the measured system states in real-time from the sensor. Meanwhile, the sensor and observer train the split Koopman autoencoder and the first phase is maintained until the autoencoder is well-trained. In the second phase, the sensor stops sending the system states, but since the split Koopman autoencoder is trained, the observer can predict the future states.


For training the split Koopman autoencoder, we consider three types of loss functions explained below.
\begin{enumerate}
    \item For accurate reconstructions of the system states, we define a \emph{reconstruction loss}, which measures the mean squared error (MSE) between the noisy system state $\hat{\mathbf{x}}_t$ and the decoded noisy representation $\Psi^{-1} (\hat{\mathbf{z}}_t)$, as
    \begin{align}
    \label{Reconstruction_Loss}
    \mathcal{L}_{\text{reconst}} = \frac{1}{T_{\scriptscriptstyle \text{P1}}} \sum_{t = 1}^{T_{ \scriptscriptstyle \text{P1}}} || \hat{ \mathbf{x}}_{t} - \Psi^{-1} \left(  \hat{\mathbf{z}}_{t} \right) ||^{2}_2,
    \end{align}
    where $T_{  \scriptscriptstyle \text{P1}}$ denotes the number of sampled system states in the first phase of the remote monitoring. Note that the sensor must consistently send the representation of the system states $\mathbf{z}_t$ along with the system states, during the first phase of the remote monitoring for computing the reconstruction loss at the observer.
   \item For obtaining the Koopman matrix $\mathbf{K}$ that guarantees linearity in the invariant subspace, we define a \emph{linearity loss}, which measures the MSE between the noisy representation of the future system state $\hat{\mathbf{z}}_{t+1}$ and multiplication of the Koopman matrix and noisy representation of the current system state $\mathbf{K}\hat{\mathbf{z}}_{t}$, as 
   \begin{align}\label{Linearity_Loss}
   \mathcal{L}_{\text{linear}} = \frac{1}{T_{ \scriptscriptstyle \text{P1}}} \sum_{t = 1}^{T_{ \scriptscriptstyle \text{P1}}} \sum_{\tau = 1}^{T_d} || \hat{ \mathbf{z}}_{t+\tau} - \mathbf{K}^{\tau} \hat{\mathbf{z}}_{t} ||^{2},
   \end{align}
   where $T_d$ is the target prediction depth considered for training.
   \item For accurate predictions of future system states, we define a \emph{prediction loss}, which measures the MSE between the noisy future system states $\hat{\mathbf{x}}_{t + \tau}$ and decoded multiplication of the Koopman matrix and noisy representation of the current system state $\Psi^{-1}(\mathbf{K}^\tau\hat{\mathbf{z}}_{t})$, as
   \begin{align}
      \label{Prediction_Loss}
      \mathcal{L}_{\text{pred}} = \frac{1}{T_{ \scriptscriptstyle \text{P1}}} \sum_{t = 1}^{T_{ \scriptscriptstyle \text{P1}}} \sum_{\tau = 1}^{T_d} || \hat{ \mathbf{x}}_{t + \tau} - \Psi^{-1} \left( \mathbf{K}^{\tau} \hat{\mathbf{z}}_{t} \right) ||^{2}.
  \end{align} 
\end{enumerate}

\begin{table}[t]
    \centering
    \resizebox{.9\columnwidth}{!}{\begin{minipage}[t]{.97\columnwidth}
    \caption{Prediction RMSE for different state representation dimension and transmission power.}
    \label{Table_1}
\begin{tabularx}{1\linewidth}{p{1.75cm} p{1.75cm} p{4cm}}
    \toprule[1pt]
    Representation Dimension $q$   & Transmission Power [W]  & Prediction RMSE [dBm] \\     
    \cmidrule(r){1-1} \cmidrule(r){2-2} \cmidrule(r){3-3}
  $1$   &  $0.1$   & $24.00$ \hspace{5pt}  \tikz{
        \fill[fill=color1] (0.0,0) rectangle (2.4,0.2);
    } \\
     & $1$  & $23.42$ \hspace{5pt}  \tikz{
        \fill[fill=color2] (0.0,0) rectangle (2.34,0.2);
    } \\
      & $10$ & $23.09$ \hspace{5pt}  \tikz{
        \fill[fill=color3] (0.0,0) rectangle (2.31,0.2);
     } \\
    &  $100$  & $\mathbf{22.94}$ \hspace{3pt}  \tikz{
        \fill[fill=color4] (0.0,0) rectangle (2.28,0.2);
    } \\ \hline
    $2$ &  $0.1$  & $15.31$\; \hspace{3pt}  \tikz{
        \fill[fill=color1] (0.0,0) rectangle (1.53,0.2);
    } \\
      &  $1$  & $12.55$\; \hspace{3pt}  \tikz{
        \fill[fill=color2] (0.0,0) rectangle (1.25,0.2);
    }  \\ 
     &  $10$  & $10.75$\; \hspace{3pt}  \tikz{
        \fill[fill=color3] (0.0,0) rectangle (1.07,0.2);
    } \\ 
    &  $100$  & $\mathbf{9.03}$ \hspace{7pt}  \tikz{
        \fill[fill=color4] (0.0,0) rectangle (0.9,0.2);
    } \\ \hline
    $3$ &  $0.1$  & $11.13$\; \hspace{3pt}  \tikz{
        \fill[fill=color1] (0.0,0) rectangle (1.11,0.2);
    } \\
      &  $1$  & $10.37$\; \hspace{3pt}  \tikz{
        \fill[fill=color2] (0.0,0) rectangle (1.03,0.2);
    } \\
     &  $10$  & $8.69$\; \hspace{7pt}  \tikz{
        \fill[fill=color3] (0.0,0) rectangle (0.87,0.2);
    } \\
     &  $100$  & $\mathbf{8.63}$ \hspace{7pt}  \tikz{
        \fill[fill=color4] (0.0,0) rectangle (0.86,0.2);
    } \\ \hline
     $4$   &  $0.1$   & $10.12$\;\hspace{5pt}  \tikz{
        \fill[fill=color1] (0.0,0) rectangle (1.01,0.2);
    } \\
      & $1$  & $9.70$\;\hspace{9pt}  \tikz{
        \fill[fill=color2] (0.0,0) rectangle (0.94,0.2);
    } \\
      & $10$ & $8.06$\;\hspace{9pt}  \tikz{
        \fill[fill=color3] (0.0,0) rectangle (0.80,0.2);
     } \\
   &  $100$  & $\mathbf{7.92}$ \hspace{6pt}  \tikz{
        \fill[fill=color4] (0.0,0) rectangle (0.79,0.2);
    } \\
    & &     \ \ \ \ \ \ \ \ \,\tikz{
        \draw[black] (1.0,0) -- (3.7,0);
        \draw[black] (1.0,-2pt) -- (1.0,2pt)node[anchor=north] {\tiny$0$};
        \draw[black] (1.8,-2pt) -- (1.8,2pt)node[anchor=north] {\tiny$8$};
        \draw[black] (2.6,-2pt) -- (2.6,2pt)node[anchor=north] {\tiny$16$};
        \draw[black] (3.4,-2pt) -- (3.4,2pt)node[anchor=north] {\tiny$24$};
    } \vspace{0mm} \\
    \bottomrule[1pt]
\end{tabularx}
 \end{minipage}}
\end{table}
 
  

The overall weighted-sum loss is defined as
\begin{align}\label{eq_Overall Loss}
\mathcal{L}_{\text{overall}} = b_{1} \mathcal{L}_{\text{reconst}} + b_{2}  \mathcal{L}_{\text{linear}} + b_{3} \mathcal{L}_{\text{pred}}
\end{align}
where $b_1,b_2,b_3 $ are positive hyperparameters. We train the model weights with stochastic gradient descent (SGD), and we consider a feedback channel from the observer to the sensor for backpropagation. An early stopping strategy is exploited to avoid model overfitting, which enhances both prediction accuracy and communication-efficiency. Once the model training is completed, the remote monitoring enters the second phase, where the system states are predicted via the split Koopman autoencoder. The prediction accuracy is evaluated using the root mean square error (RMSE) given as
\begin{align}\label{eq:RMSE}
\text{RMSE} = \sqrt{ \frac{1}{T_{\scriptscriptstyle \text{P2}}} \sum_{ t  =T_{\scriptscriptstyle \text{P1}+1} }^{T_{\scriptscriptstyle \text{P1}} + T_{\scriptscriptstyle \text{P2}}} \left( \bar{\mathbf{x}}_{t} - \hat{\mathbf{x}}_{t} \right)^2 },
\end{align}
where $T_{\scriptscriptstyle \text{P2}}$ is the number of predictions that the observer makes in the second phase of the remote monitoring. Note that if the second phase becomes longer, there will be error propagation that will degenerate the performance of the prediction. Then, we have two options to enhance the performance: 1) sending a new system state representation to initialize the prediction, and 2) shifting to the first phase of remote monitoring to fine-tune the split Koopman operator. The effectiveness of the proposed wireless split Koopman autoencoder is validated in the following section. \\

\begin{table}[t]
    \centering
    \resizebox{0.9\columnwidth}{!}{\begin{minipage}[t]{.97\columnwidth}
    \caption{Prediction RMSE for different state representation dimensions and training periods.}
    \label{Table_2}
\begin{tabularx}{1\linewidth}{p{1.7cm} p{2.2cm} p{3.6cm}}
    \toprule[1pt]
     Representation Dimension $q$ & Training Period [s] & Prediction RMSE [dBm]\\     
    \cmidrule(r){1-1} \cmidrule(r){2-2} \cmidrule(r){3-3}
  $1$   &  $150$   & $26.72$\hspace{5pt}  \tikz{
        \fill[fill=color1] (0.0,0) rectangle (2.67,0.2);
    } \\
     & $250$  & $26.02$\hspace{5pt}  \tikz{
        \fill[fill=color2] (0.0,0) rectangle (2.60,0.2);
    } \\
      & $350$& $\textbf{23.09}$\hspace{7pt}  \tikz{
        \fill[fill=color3] (0.0,0) rectangle (2.31,0.2);
     } \\ \hline
    $2$ &  $150$  & $18.26$ \hspace{3pt}  \tikz{
        \fill[fill=color1] (0.0,0) rectangle (1.82,0.2);
    } \\
       &  $250$  & $16.03$ \hspace{3pt}  \tikz{
        \fill[fill=color2] (0.0,0) rectangle (1.60,0.2);
    } \\
     &  $350$  & $\textbf{10.75}$ \hspace{5pt}  \tikz{
        \fill[fill=color3] (0.0,0) rectangle (1.07,0.2);
    } \\ \hline
   $3$ &  $150$  & $15.36$ \hspace{3pt}  \tikz{
        \fill[fill=color1] (0.0,0) rectangle (1.53,0.2);
    } \\
     &  $250$  & $12.78$ \hspace{3pt}  \tikz{
        \fill[fill=color2] (0.0,0) rectangle (1.27,0.2);
    } \\
      &  $350$  & $\textbf{8.69}$ \hspace{8pt}  \tikz{
        \fill[fill=color3] (0.0,0) rectangle (0.86,0.2);
    } \\ \hline
     $4$ &  $150$  & $14.12$ \hspace{3pt}  \tikz{
        \fill[fill=color1] (0.0,0) rectangle (1.41,0.2);
    } \\
      &  $250$  & $10.00$ \hspace{3pt}  \tikz{
        \fill[fill=color2] (0.0,0) rectangle (1.00,0.2);
    } \\
      &  $350$  & $\textbf{8.06}$ \hspace{8pt}  \tikz{
        \fill[fill=color3] (0.0,0) rectangle (0.80,0.2);
    } \\
    
    & & \ \ \ \ \ \ \ \ \,\tikz{
        \draw[black] (1.0,0) -- (3.7,0);
        \draw[black] (1.0,-2pt) -- (1.0,2pt)node[anchor=north] {\tiny$0$};
        \draw[black] (1.8,-2pt) -- (1.8,2pt)node[anchor=north] {\tiny$8$};
        \draw[black] (2.6,-2pt) -- (2.6,2pt)node[anchor=north] {\tiny$16$};
        \draw[black] (3.4,-2pt) -- (3.4,2pt)node[anchor=north] {\tiny$24$};
    } \vspace{0mm} \\
    \bottomrule[1pt]
\end{tabularx}
 \end{minipage}}
\end{table}

\section{Simulation Results}\label{Numerical_Results}
 In this section, we investigate the performance of the proposed split learning of Koopman autoencoder over wireless channel in the inverted cart-pole system~\cite{morasso2019stabilization}. This system is non-linear and multi-dimensional. For a given time, the system state is given by a four-dimensional vector $\mathbf{x} = \left[x, \, v, \, \theta,  \, \omega \right]$ where $x$ and $v$ denote the horizontal position and velocity of the cart, respectively. The terms $\theta$ and $\omega$ are the vertical angle and angular velocity of the pendulum, respectively. Accordingly, the system dynamics described as follows.
 
 \small
 \begin{align}\label{CIP_Dynamics}
 \begin{split}
\frac{dx}{dt} &= v,\\
\frac{d \theta}{dt}  &= \omega \\
\frac{dv}{dt} &= \frac{-m^{2} L^{2} g \cos(\theta) \sin(\theta) + m L^{2} (m L\omega^{2} \sin(\theta) - \delta v )  }{m L^{2} \left(M + m (1 - \cos(\theta)^{2} ) \right)  }\\
\frac{d \omega}{dt} & = \frac{ (m + M) mg L \sin(\theta) - m L \cos(\theta) ( m L\omega^{2} \sin(\theta) - \delta v )  }{ m L^{2} \left( M + m ( 1 - \cos(\theta)^{2} )\right)}
 \end{split}
 \end{align}\normalsize

\noindent Here, we consider the following simulation parameters: pendulum mass $m = 1$ Kg, cart mass $M = 5$ Kg, pendulum length $L = 0.2$ m, gravitational acceleration $g = -10$  $\text{m/} \text{s}^{2}$, and the cart damping $\delta = 1 \text{N s} /\text{m}$. 
The training dataset is generated based on the received noisy states from the inverted cart-pole system with a sampling rate of $10$ ms in the time interval $t \in [0, |\mathcal{T}_{ \text{Train}}|]$ and an initial state of $\mathbf{x} = \left[0, \; 0, \;  3.14, \; -0.5 \right]$. The split Koopman autoencoder weights are trained to minimize the overall unweighted-sum loss function in \eqref{eq_Overall Loss} via the Adam optimizer~\cite{sutskever2013importance} with batch size of $128$. The training is continued for $20$ training epochs. The encoder of the Koopman autoencoder consists of four fully-connected layers that contain $128$, $64$, $32$ and $q$ neurons with rectified linear unit (ReLu) activation, respectively. The decoder begins with $2$ fully-connected layers with the width $q\in\{1, 2, 3, 4 \} $ (i.e., encoded state representation dimension), and the rest follows the same structure of the encoder. The wireless communication channels between the encoder and decoder follow Rayleigh fading channels. We consider a transmission power $ P \in \{0.1, 1, 10, 100 \} \; \text{Watts}$, the distance between the sensor and the observer $R = 1$ Km, and the path loss exponent $\alpha = 2$. 



\begin{figure}[t]
    \centering
    \subfigure[Cart Position.]{\includegraphics[width=0.24\textwidth]{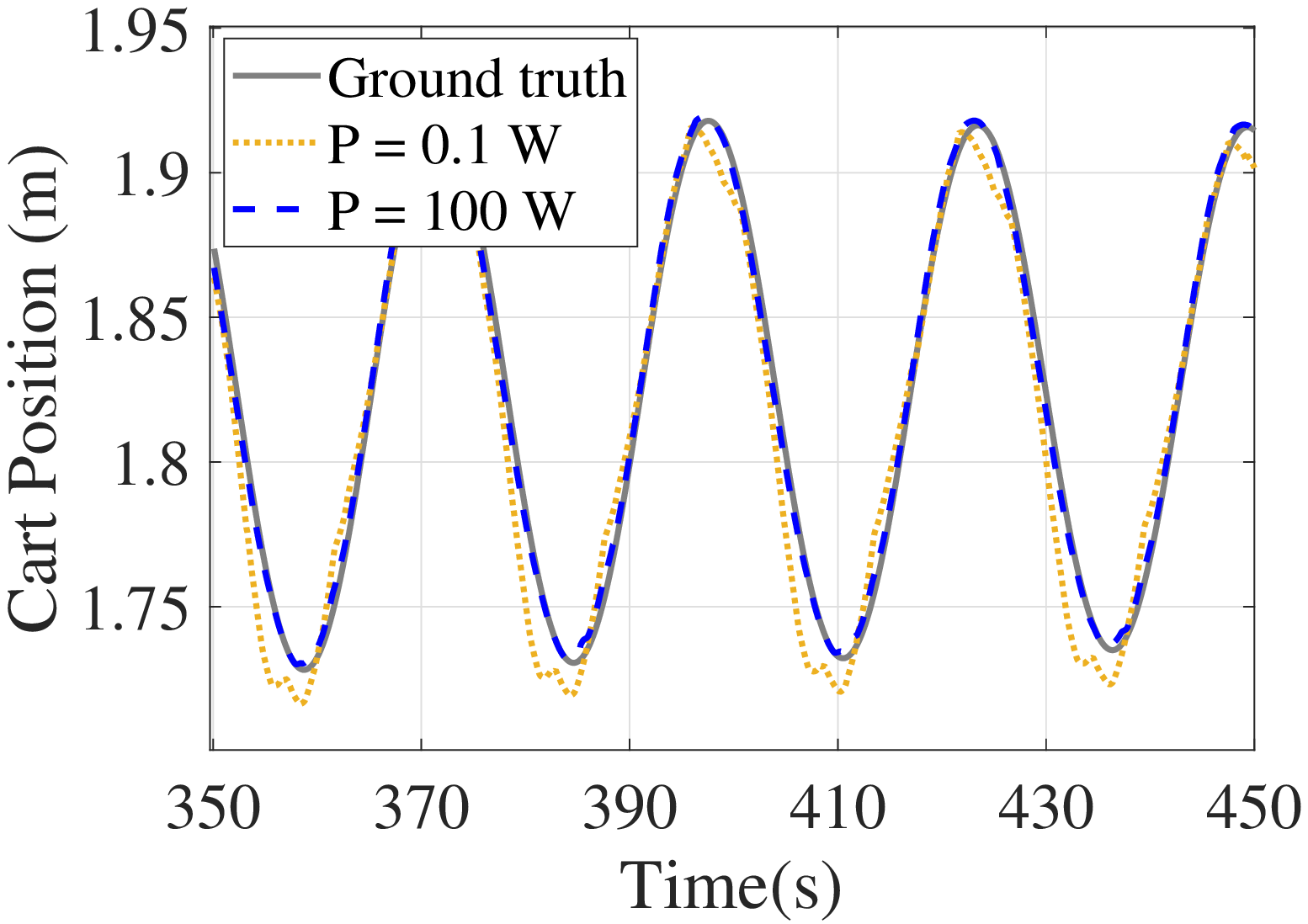}} 
    \subfigure[Cart Velocity.]{\includegraphics[width=0.24\textwidth]{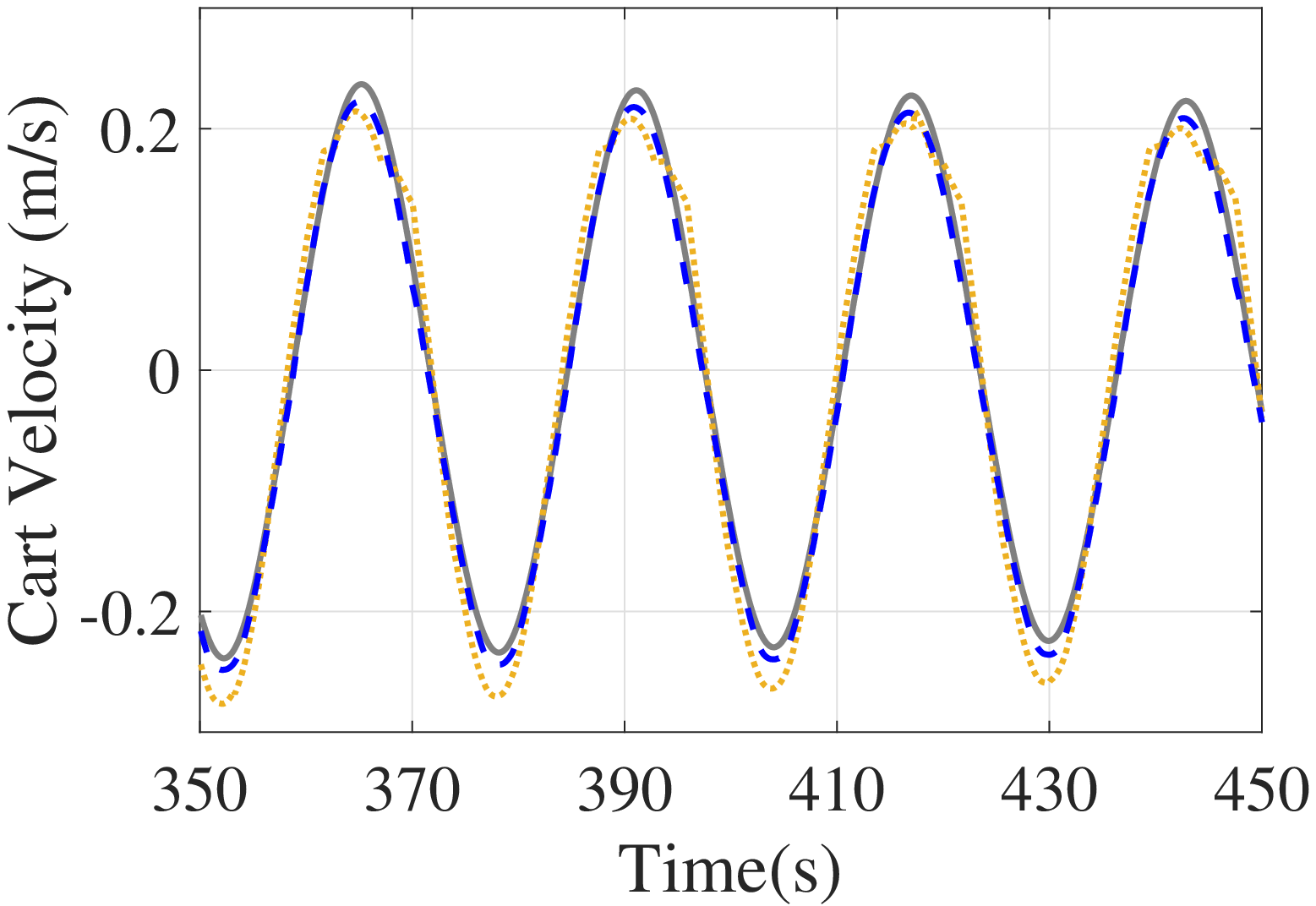}} 
    \subfigure[Pendulum Angular Position.]{\includegraphics[width=0.24\textwidth]{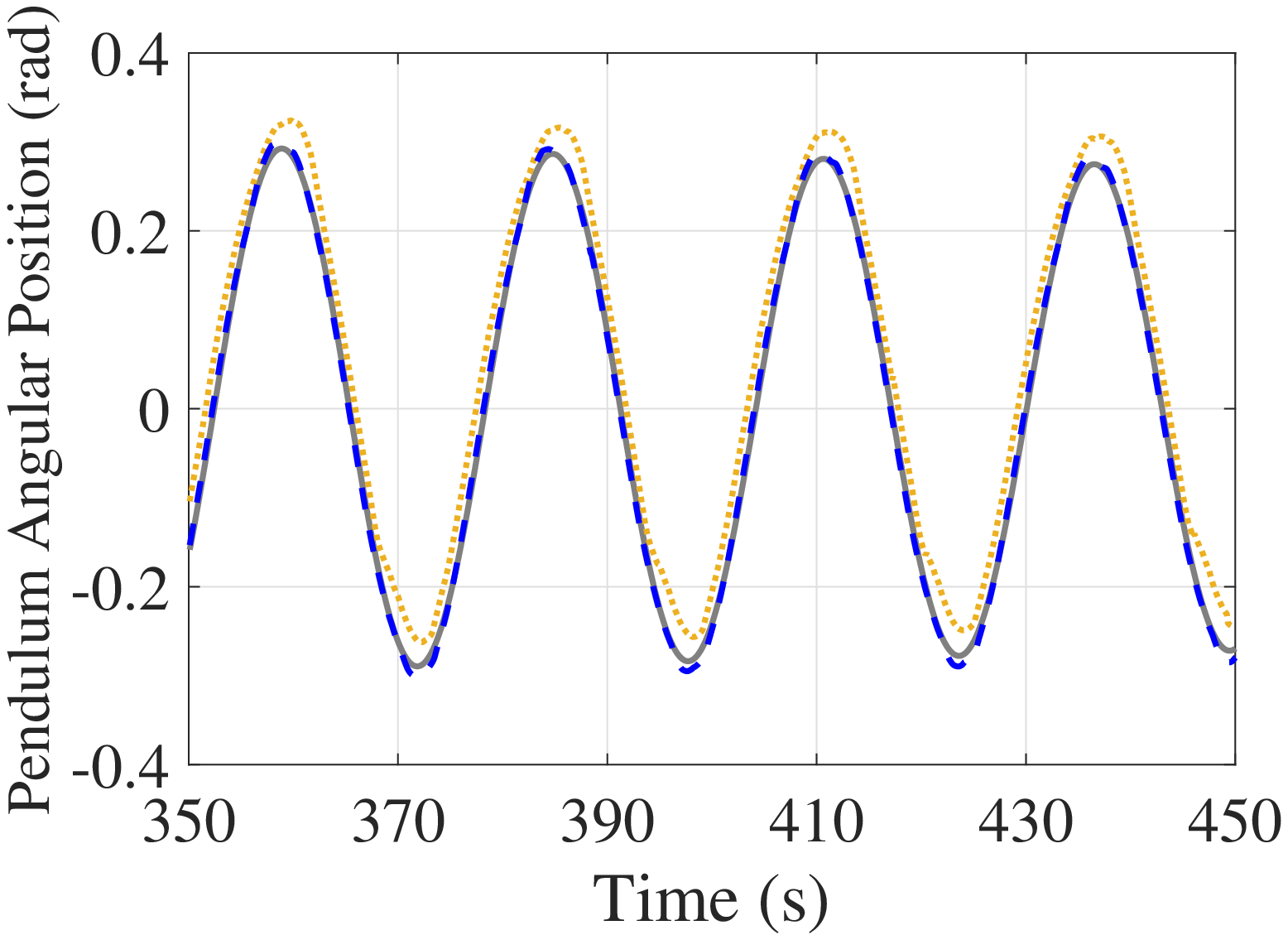}}
    \subfigure[Pendulum Angular Velocity.]{\includegraphics[width=0.24\textwidth]{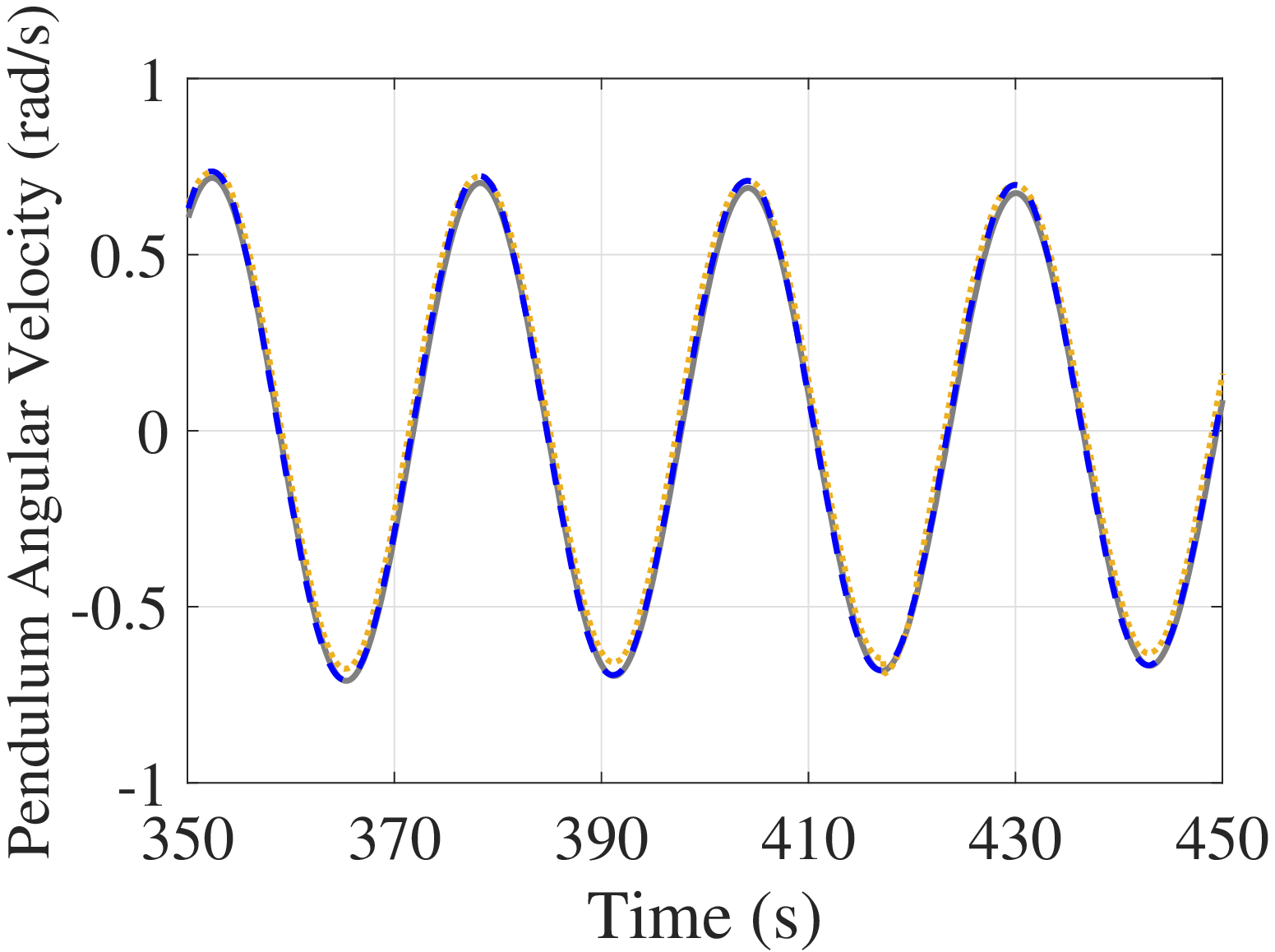}}
    \caption{Time-series of the predicted inverted cart-pole system states utilizing the proposed split Koopman autoencoder before observing the ground truth.}
    \label{fig state_prediction}
\end{figure}

\vspace{3pt}\noindent\textbf{Tx Power vs. Prediction Accuracy}.\quad Fig.~\ref{fig state_prediction} demonstrates the prediction performance of the proposed wireless split Koopman autoencoder compared to the vanilla remote monitoring of the non-linear inverted cart-pole system. The remote observer in the proposed wireless split Koopman autoencoder predicts the future time-series non-linear system states for different transmission power $P \in \{ 0.1, 100 \} \; \text{Watts}$  compared to the remote observer in the vanilla remote monitoring that observes the actual system states over an ideal channel. It can be seen that the predicted system states in the proposed wireless split Koopman autoencoder with $P = 100 \; \text{Watts}$ match closely to the observed system states in the vanilla remote monitoring up to $1000$ discrete time-steps compared to the predicted system states with $P = 0.1 \; \text{Watts}$. 

 The rationale behind this result is that the stability of learning the wireless split Koopman autoencoder depends on the communication reliability of the received state observations at the remote observer that is affected by the transmission power, highlighting the importance of jointly design the communication and deep learning operations. As a consequence, the prediction accuracy at the remote observer can be improved at the cost of increasing the communication cost in terms of the transmission power. Furthermore, the prediction results in Fig.~\ref{fig state_prediction} corroborate the ability of the wireless split Koopman autoencoder to discover the Koopman invariant subspace with two-dimensional encoded state representations. From this we can see that the proposed wireless split Koopman autoencoder is able to predict the future system states based on the trained Koopman matrix.

\vspace{3pt}\noindent\textbf{Tx Power \& Representation Dim. vs. Prediction Accuracy}.\quad As observed in the prediction results in Fig.~\ref{fig state_prediction}, the proposed wireless split Koopman autoencoder performs better under high transmission power and the two-dimensional encoded state representations. Table~\ref{Table_1} demonstrates the prediction accuracy for different transmission power and different encoded state representation dimensions. It can be seen that the prediction accuracy is improved at the cost of increasing transmission power for the same encoded state representation dimensions and the same training period, while the prediction accuracy is almost the same in low and high encoded state representation dimensions, i.e., $q \in \{ 1,4 \}$, for different values of transmission power. In contrast, increasing the transmission power has a notable effect in increasing the prediction accuracy in the other encoded state representation dimensions $q \in \{ 2,3 \} $ compared to the other dimensions. 
 
 The reason behind the results in Table~\ref{Table_1} is that $q = 1$ is not sufficient to discover the Koopman invariant subspace, while $q = 4$ has more redundant information than the Koopman invariant subspace. Hence, selecting a high transmission power and large dimensions of encoded state representations is instrumental in casting the non-linear system dynamics into a linear framework at the cost of increasing the communication cost in terms of
 transmission power and communication payload size, leading to a trade-off between communication cost and prediction accuracy.

 
  \begin{figure}[t]
    \centering
    \subfigure[]{\includegraphics[width=0.24\textwidth]{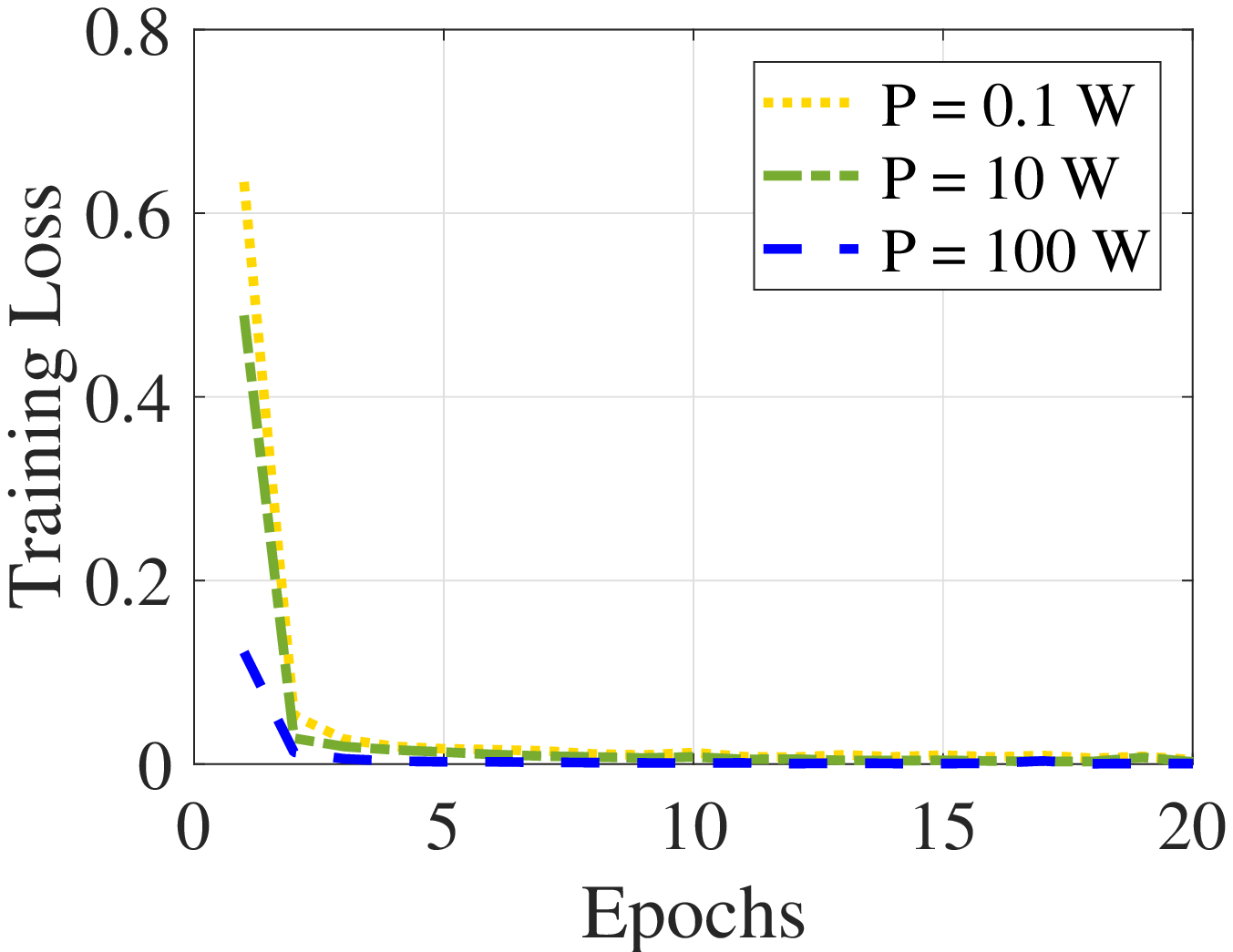}} 
    \subfigure[]{\includegraphics[width=0.24\textwidth]{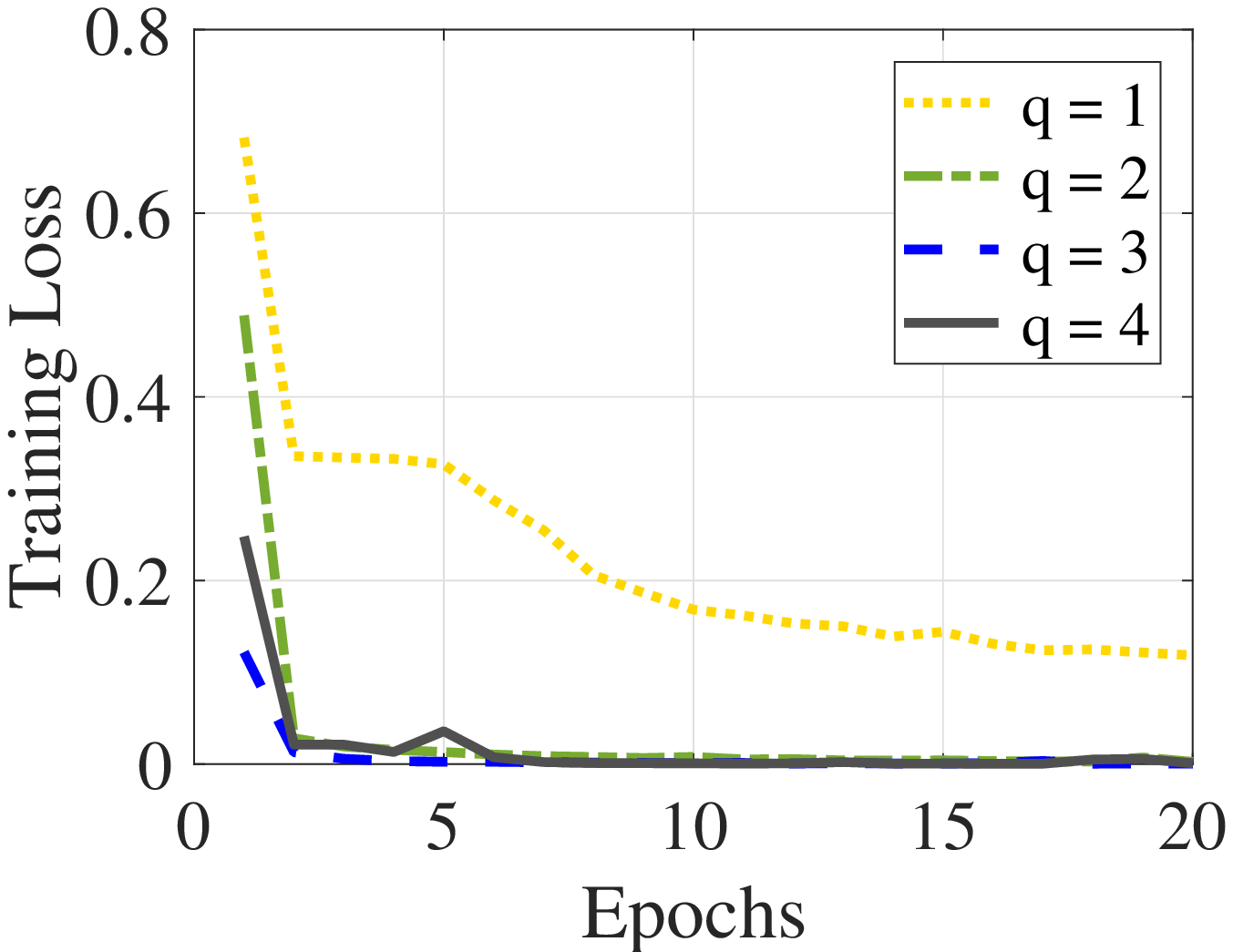}} 
    \caption{Training loss of the proposed split Koopman learning with (a) different transmission power and (b) different encoded state representation dimensions. }
    \label{fig Train_Loss}
\end{figure}

 \vspace{3pt}\noindent\textbf{Training Period \& Representation Dim. vs. Prediction~Accuracy}.\quad As opposed to the vanilla remote monitoring wherein the remote observer receives the actual system states each time due to the difficulty in identifying the non-linear system dynamics, our proposed wireless split Koopman autoencoder predicts the future system states even without communication after observing a sufficient number of sampled system states, highlighting the effectiveness of the wireless split Koopman autoencoder in enabling communication-efficient identification and prediction of the non-linear system states. Table~\ref{Table_2} demonstrates the prediction accuracy for the same transmission power $P = 10 \; \text{Watts}$, different training periods and different encoded state representation dimensions. The prediction accuracy is improved as the training period increases for the same transmission power and same encoded state representation dimension, highlighting the trade-off between the prediction accuracy and the communication payload size. The prediction accuracy in Table~\ref{Table_2} is improved as a result of increasing the number of received training state observations by increasing the training period for the same transmission power and the same encoded state representation dimension. 
 
 \vspace{3pt}\noindent\textbf{Tx Power \& Representation Dim. vs. Training Loss}.\quad Fig.~\ref{fig Train_Loss} shows the training loss of the proposed wireless split Koopman autoencoder for different transmission power and different encoded state representation dimensions. It is clear that the training loss generally drops as the number of epochs increases. Moreover, the increment of the transmission power or encoded state representation dimension fasten the convergence of the split Koopman autoencoder training, thereby reducing the overall communication payload size thanks to the short training time.

\section{Conclusion}
\label{conclusion}
In this paper, we proposed a novel wireless split Koopman autoencoder for linearizing non-linear dynamical systems, thereby reducing the communication payload size for remote monitoring and enabling the prediction of the future states of the dynamical system. Throughout the inverted cart-pole control simulations, we demonstrated the effectiveness of the proposed method in future state prediction with high accuracy for practical ranges of transmission power, state representation dimensions, and training periods. Extending this single-sensor and open-loop scenario to multi-sensor scheduling and closed-loop control scenarios could be an interesting topic for future research.





\bibliographystyle{IEEEtran}
\bibliography{IEEEabrv,bibliography}
\end{document}